# Application of multimodal fusion deep learning model in disease recognition

Xiaoyi Liu[1], Hongjie Qiu[2], Muqing Li[3], Zhou Yu[4], Yutian Yang[5], Yafeng Yan[6]

[1]Arizona State University, USA, xliu472@asu.edu

[2]University of Washington, USA, hongjieq@uw.edu

[3]University of California San Diego, USA, MUL003@ucsd.edu

[4]University of Illinois at Chicago, USA, zyu941112@gmail.com

[5]University of California, Davis, USA, yytyang@ucdavis.edu

[6]Stevens Institute of Technology, USA, yanyafeng0105@gmail.com

*Abstract*—This paper introduces an innovative multi-modal fusion deep learning approach to overcome the drawbacks of traditional single-modal recognition techniques. These drawbacks include incomplete information and limited diagnostic accuracy. During the feature extraction stage, cutting-edge deep learning models including convolutional neural networks (CNN), recurrent neural networks (RNN), and transformers are applied to distill advanced features from image-based, temporal, and structured data sources. The fusion strategy component seeks to determine the optimal fusion mode tailored to the specific disease recognition task. In the experimental section, a comparison is made between the performance of the proposed multi-mode fusion model and existing single-mode recognition methods. The findings demonstrate significant advantages of the multimodal fusion model across multiple evaluation metrics.

## I. INTRODUCE

Application of Multi-modal Fusion Deep Learning Model in Disease Recognition" is a cutting-edge research that focuses on how to improve the accuracy and efficiency of disease recognition in the medical field by integrating multiple types of data information and using advanced deep learning technology. With the rapid development of medical data acquisition technology, the multi-modal characteristics of medical information have become increasingly prominent from traditional medical imaging to genomic data and electronic medical records [1-3]. However, how to effectively integrate these heterogeneous data sources and tap their internal connections to support more accurate disease diagnosis has become one of the core challenges of current medical artificial intelligence research.

Although the traditional disease identification method based on a single mode has achieved remarkable results in some specific fields, its diagnostic accuracy and comprehensiveness are often limited[4]. Single modal data is often difficult to fully reflect the complexity of the disease, especially in the face of early lesions, and heterogeneous diseases, the uncertainty of diagnosis is significantly increased. Therefore, exploring new ways of multimodal data fusion has become the key to improve the efficiency of disease identification.

Multimodal data mainly includes but is not limited to medical image data (such as X-rays, CT, MRI), physiological signals (electrocardiogram, electroencephalogram), genomic data, clinical laboratory test results, and electronic medical record records [5-7]. Each modal data provides information about the disease state from a different perspective, and the purpose of multimodal fusion is to integrate this complementary information to form a more comprehensive view of the disease.

Firstly, according to the characteristics of different modal data, a special deep learning model is designed for feature extraction. For instance, Convolutional Neural Networks (CNNs) excel in learning pixel-level features from image data, adept at capturing structural organization and anomalies [8-10]. RNN (Recurrent Neural Networks) or LSTM (Long Short-Term Memory networks) are well-suited for processing time series data like electrocardiogram signals, effectively capturing temporal dynamics [11]. In cases of structured data such as clinical indicators, Fully Connected Networks (FCNs) serve for feature encoding purposes.

Multi-modal data fusion strategy is the core of the research. In the early stage of fusion, the input layer directly merges the data of different modes to promote mutual learning between modes[12-13]. In the late fusion, after the learning of each modal feature is completed, the combination is carried out in the high level feature space to maintain the independence of the modal characteristics[14-15]. Intermediate fusion is a feature fusion at multiple levels, trying to combine the advantages of both. Through comparative analysis, this study explored the optimal fusion strategy suitable for specific disease recognition tasks.

In order to verify the validity of the proposed model, a comprehensive multimodal dataset was constructed by collecting a large number of imaging data, clinical indicators and genetic information. The stability and effectiveness of the model training are ensured by fine data preprocessing. The experimental results show that the multi-modal fusion deep learning model can significantly improve the recognition accuracy, especially in identifying early lesions and reducing the misdiagnosis rate.

This paper not only demonstrates the great potential of deep learning and multi-modal data fusion in improving disease recognition ability, but also provides new ideas for future medical artificial intelligence research. By continuously optimizing the model structure, exploring more efficient integration strategies, and strengthening the explainability and ethical norms of the model, this research direction is expected to further promote the development of precision medicine and provide patients with more timely and accurate diagnostic services. In the future, with the enhancement of computing power and the continuous progress of algorithms, the multi-modal deep learning model will show its unique value in more disease fields and open a new chapter of intelligent healthcare.

## II. RELATED WORK

With the continuous evolution of deep learning technology, convolutional neural network (CNN) has demonstrated its power in many image fusion fields, covering complex scenes such as multi-focus fusion, multi-exposure fusion, visible and infrared spectral fusion, and medical image fusion [16-19]. For example, the innovative scheme introduced by Ma et al., an adversarial generation network equipped with dual discriminators, can efficiently complete the task of cross-resolution image fusion without damaging thermal radiation information and visible texture details [20].

Acosta et al. implemented quantitative optimization of fusion image quality by incorporating structural similarity index into training loss function. This method further promoted clear visualization of fusion image by evaluating the contribution of each input image to the final fusion result [21]. It is worth noting that some studies have proposed an unsupervised and unified dense connected network model[22], which uses data-driven weight strategies to automatically adjust the importance of features of each source image, and constructs an efficient model to meet diverse fusion requirements through an elastic weight combining mechanism.

Zhang et al. 's work focused on the end-to-end full convolution fusion framework, which captures the significant features of images with the help of the convolution layer, and customates fusion strategies according to the image characteristics, and finally outputs information-rich fusion images through the reconstruction process . In addition, the fusion network designed by Gao et al. incorporates nested connections and spatial/channel attention mechanisms. This design not only preserves image information at multiple scales, but also improves the fusion effect by accurately locating key areas and features through attention allocation at the spatial and channel levels[23].

Although deep learning-driven image fusion technology often surpasses traditional methods in terms of performance and flexibility, it faces several inherent challenges[24]. These include dependence on large-scale labeled datasets, limitations in model generalization, and high consumption of computing resources during training, such as a large memory footprint and extended training cycles. However, recent findings by Yan et al. highlight significant strides in addressing these resource-intensive demands within medical contexts. Their research demonstrates how advancements in neural network efficiency can considerably mitigate these challenges, particularly in applications like survival prediction across diverse cancer types[25]. This progress lays a strong foundation for the ongoing development of more efficient, generalizable, and resource-friendly deep learning fusion strategies. Such innovations are crucial for advancing the field of medical disease recognition, making it not only more effective but also more accessible, thus paving the way for broader adoption and implementation in healthcare settings.

## III. THEORETICAL BASIS

### A. Convolutional neural network

Convolutional Neural Networks (CNNs), pioneered by Fukushima, stand as pivotal models within deep learning, notably prominent in computer vision for image processing tasks. Unlike traditional algorithms necessitating extensive data preprocessing for intricate datasets, CNNs excel in automatic feature extraction. This capability positions them as a mainstream approach for data analysis, owing to their inherent capacity for automatic feature learning. CNN is a machine learning algorithm that guides the model to perform tasks such as classification and segmentation by finding mapping relationships in data, and its essence is a relatively complex function.

The convolutional layer serves the purpose of extracting feature representations from image data. The pooling layer is used for dimensionality reduction of features to reduce the number of parameters in subsequent networks: the fully connected layer is used for linear combination of final features. After the convolution and pooling operations, the neural network obtains the ability to fit real-world samples through the mapping of the activation functions. After multiple sequential superposition of convolution - pooling modules, the neural network uses one or more fully connected layers to extract feature maps and randomly combine them to get the best features in the final classification, so as to obtain the output result of the network. In addition to the convolution operation, another feature of CNN is weight sharing. In the same convolutional layer, all neurons share the same weight, which makes CNNS have good locality and shift invariance when processing images. In addition, due to the existence of weight sharing mechanism, CNN can reduce the risk of overfitting to a certain extent when faced with fewer parameters. The CNN architecture is shown in Figure 1.

The convolution layer in CNN serves as the central element for feature extraction. It processes input images through convolution operations, generating convolutional feature maps. By utilizing convolution kernels of various sizes, distinct features can be extracted. These convolution kernels' parameters are learned during training, with the backpropagation algorithm adjusting them to enhance the layer's feature extraction abilities. Convolution operations entail sliding convolutional filters across input data, resulting in convolutional feature maps that represent specific input data features. The input and output dimensions of the convolutional layer are defined as $W_{in} * H_{in} * D_{in}$ and $W_{out} * H_{out} * D_{out}$ respectively.

$$W_{out} = \frac{W_{in} - F + 2P}{S} + 1 \quad (1)$$

$$H_{out} = \frac{H_{in} - F + 2P}{S} + 1 \quad (2)$$

In the context provided, $F$ denotes the size of the designed convolution kernel, specifying its dimensions. $S$ represents the step length, determining the number of pixels traversed between successive convolution operations on the feature map. $P$ signifies the zero-padding operation, which involves adding zero values around the input data to adjust the dimensions for convolution operations.

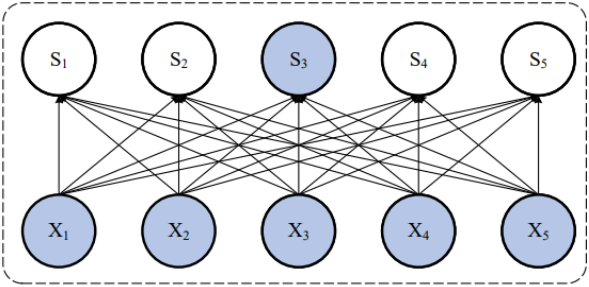

Fig. 1. Convolutional neural network

*B. Recurrent neural network*

The recurrent neural network model was proposed in 1943 by McCulloch and Peters et al., as shown in Figure 2. In their masterpiece, McCulloch, Peters and others proposed a mathematical model that can be used to process information by neural networks, and proved that it can be used to solve any function with an algorithm. Since this time, neural networks and human neural networks have emerged and are developing rapidly. Neural network is a kind of spiritual network with learning ability, and it can grasp the internal law of relatively complex things. In a neural network model, neurons process input information, and this information is the basic condition for the operation of the neural network. Neurons are the basic basis of neural network design. The specific formula can be as follows:

$$u_k = \sum_{j=1}^{m} w_{kj} x_j \quad (3)$$

$$y_k = \varphi(u_k + b_k) \quad (4)$$

In the above formula, $w_{kj}$ is the neuron weight, $x_j$ is the input value, $u_k$ is the output value during each iteration, and $y_k$ is the final output value

In the process of multiple iterations, the offset item $bk$ will affect the most important item of the final output value, which belongs to the external parameter of the neuron and is responsible for adjusting the output value each time. Therefore, the final output value of the neuron $y_k$ is:

$$y_k = \sum_{j=0}^{m} wkjXj \quad (5)$$

$$y_k = \varphi(v_k) \quad (6)$$

Recurrent neural network is a kind of deep learning neural network with strong nonlinear characteristics, which is widely used in image recognition, signal processing, automatic control and other fields, and plays an important role in data mining, rule discovery and fault diagnosis.

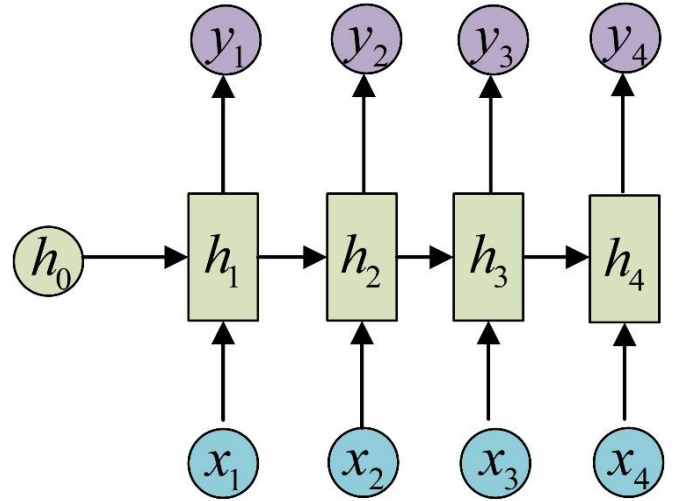

Fig. 2. Recurrent neural network framework

*C. Transformer*

In 2017, Vaswani et al. proposed Transformer model in Natural Language Processing (NLP) field. In addition, Vaswani's Transformer essentially inputs a sequence and outputs a sequence model (sequence to sequence). Inspired by Transformer, Dosovitskiy et al. introduced Transformer architecture into the field of computer Vision and designed Vision Transformer according to the data characteristics in computer vision. In addition, many researchers have been inspired by this work to apply vision Transformer to their own fields. In the field of vision, Transformer has achieved superior performance in a number of typical tasks of computer vision. The structure of the Vision Transformer model includes image serialization, linear transformation, position coding, Transformer encoder and a classifier composed of a fully connected network, as shown in Figure 3.

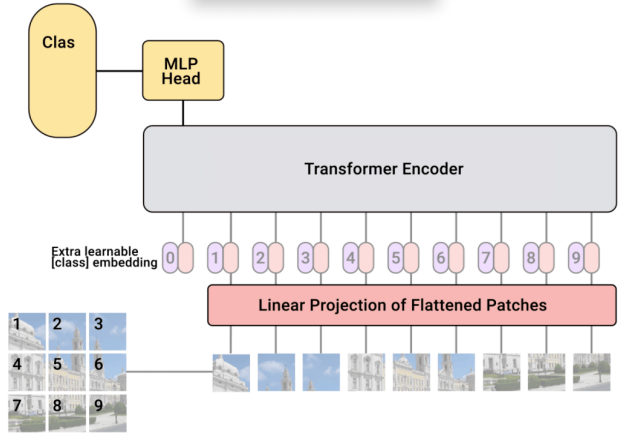

Fig. 3. Vision transform framework

In essence, the input and output of Transformer are serial models, so the design idea of Vision Transformer also needs to convert images into serial forms. As shown in Figure 3, the Vision Transformer model needs to first divide the image into 9 equally large regions and small images, and then modify the image size of the divided image through linear transformation. Since the location information of each part is not saved in the process of image segmentation, in order to make the model have location features in the subsequent training and feature extraction process. Therefore, position coding is added to each part of the image after linear transformation. Finally, all parts of the image data after adding the position coding are spliced back and forth, and the whole data is sent into the subsequent structure of the model as a sequence for calculation.

The reason why Vision Transformer can achieve excellent results in the field of computer vision is mainly because it uses the attention mechanism[26], and the attention mechanism is different from the feature extraction method of convolutional neural network. The scaling dot-product attention mechanism is adopted in the model design of Vision Transformer.

$$Attention = softmax(\frac{QK^T}{\sqrt{d_q}})V \qquad (7)$$

In the formula, Q, K and V represent Q(Queries), K (Keys) and V (Values) respectively, Attention represents the calculated attention score, and pdq represents the dimensionality of matrix Q. Q, K, and V can be calculated by the following formula.

$$Q = W_q X \qquad (8)$$

$$K = W_k X \qquad (9)$$

$$V = W_v X \qquad (10)$$

## IV. MODEL

This paper constructs an innovative model that integrates the advantages of c CNN, RNN, and Vision Transformer (ViT) for efficient analysis of medical image and clinical text data. The process is as follows: First, the image data is preprocessed and patch partition is implemented for ViT to capture global features; After clinical text is cleaned and word embedded, time series information is captured by RNN. CNN focuses on spatial feature extraction of images. Subsequently, the feature vectors generated by these independent modules are integrated at the multimodal fusion layer by means of attention mechanism or direct splicing to promote information complementarity. The fusion features are fed into the fully connected layer and converted into the predicted probabilities of various diseases by softmax function. The model was optimized by cross-entropy loss function, trained with early stop and learning rate adjustment strategies, and finally evaluated its accuracy and sensitivity in the test set, demonstrating the potential of cross-modal information fusion in improving the accuracy of disease recognition.

## V. EXPERIMENT

### A. Data set

Medical image data set: We used publicly available medical Imaging databases, such as lung Cancer CT scans from TCIA (The Cancer Imaging Archive), And MRI images from ADNI database for Alzheimer's disease identification studies. These datasets not only contain a large number of normal and pathological samples, but also provide detailed clinical annotations and pathological reports, providing a solid foundation for model training and validation.

Clinical text dataset: To integrate clinical text Information, we used the MIA-III database, which contains a wealth of electronic medical records covering a wide range of clinical indicators, laboratory test results, and doctors' diagnostic notes. Through the processing and analysis of these text data, we can extract key clinical features for auxiliary image data analysis.

### B. Experimental setup

Data preprocessing: All medical image data undergo a uniform preprocessing step, including standardization, normalization, and necessary image enhancement (such as rotation, flipping, scaling, etc.) to increase the generalization ability of the model. The dataset utilizes Linked Data techniques to amalgamate disparate data formats[27], thereby improving data interoperability and analytical capabilities in domains such as artificial intelligence and machine learning. This method aids in dismantling data silos, augments the variety of datasets, and fosters the creation of more precise and expansive AI applications. The image processed by ViT is divided into patches of fixed size[28]. The clinical text is cleaned, the noise information is removed, and the word embedding technique is used to transform it into numerical vectors, which are RNN-friendly. Fill or truncate the text sequence to ensure that the sequence length is consistent.

Data set partitioning: Data sets are randomly divided into training sets (70%), validation sets (15%), and test sets (15%), ensuring fairness in model training, tuning, and final performance evaluation.

Model building: The built model consists of separate CNN, RNN, and ViT modules, each responsible for processing images, text, and other data modes. Subsequently, the modal features are integrated through a designed fusion mechanism, such as attention-mechanism fusion or weighted summation.

Parameter configuration for the experiment: The Adam optimizer was utilized with an initial learning rate set at 0.001, incorporating a learning rate decay strategy to prevent overfitting during training. A batch size of 32 was chosen to optimize the use of computational resources and facilitate timely model convergence. The duration of the training was contingent on the outcomes observed in the validation set, employing an early stopping mechanism where training ceases if there is no improvement in the validation set performance after several iterations.

*C. Result analysis*

TABLE I. THE RESULTS OF DIFFERENT MODELS ON EVALUATION INDICATORS

| Method | Precision | Accuracy | F1 | Recall |
|---|---|---|---|---|
| CNN | 0.62 | 0.65 | 0.53 | 0.71 |
| RNN | 0.66 | 0.69 | 0.58 | 0.72 |
| Transformer | 0.69 | 0.73 | 0.62 | 0.75 |
| Our | 0.76 | 0.77 | 0.72 | 0.83 |

Table I presents a comparative analysis which demonstrates that the multi-modal fusion approach discussed in this study delivers superior results in disease recognition tasks. It records an Accuracy of 0.77, alongside Precision and Recall rates of 0.76 and 0.83, respectively, and achieves an F1 score of 0.72. This model outperforms the single-mode CNN (Accuracy 0.65, F1 0.53), RNN (Accuracy 0.69, F1 0.58), and Transformer (Accuracy 0.73, F1 0.62), indicating significant enhancements in both accuracy and recall capabilities. The data strongly supports the efficacy and advantage of employing a multimodal fusion strategy.

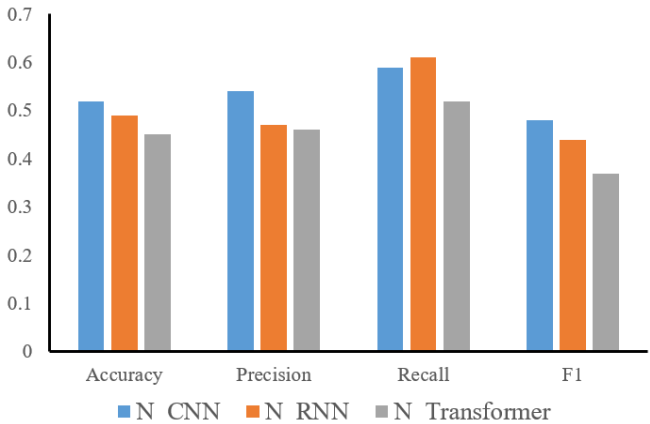

Fig. 4. Ablation results

In Figure 4, we removed different parts of the three models (N_CNN, N_RNN, N_Transformer) and observed changes in accuracy, accuracy, recall, and F1 scores. Removing key modules from the CNN, RNN, and Transformer models impacts performance to varying degrees. After removing N_CNN, the model has the least impact, after removing N_RNN, the model slightly decreases, and after removing N_Transformer, the model performance decreases sharply, indicating that Transformer module plays a key role in task execution.

## VI. CONCLUSION

In summary, aiming at the limitations of traditional single-mode disease recognition technology, such as incomplete diagnosis and limited accuracy caused by information loss, this study innovatively designed and implemented a multi-mode fusion deep learning model. The core of the model is its carefully constructed fusion framework, which successfully integrates deep learning technologies such as CNN, RNN and Transformer to achieve efficient feature extraction of medical images, time series data and structured information, which greatly enriches the depth and breadth of the model's understanding of disease characterization